\documentclass{article}

\PassOptionsToPackage{numbers, compress}{natbib}

\usepackage[final]{semilattice}

\usepackage[utf8]{inputenc}   
\usepackage[T1]{fontenc}      
\usepackage{hyperref}         
\usepackage{url}              
\usepackage{doi}             
\usepackage{booktabs}         
\usepackage{multirow}         
\usepackage{amsfonts}         
\usepackage{amsmath}          
\usepackage{nicefrac}         
\usepackage{microtype}        
\usepackage{graphicx}         
\usepackage{subcaption}       
\usepackage{float}            
\usepackage{tikz}             
\usetikzlibrary{arrows.meta, positioning, fit, backgrounds, calc}
\usepackage{xcolor}           
\usepackage{cleveref}         

\graphicspath{{figures/}}

\title{Mind the Gaps: Mixture-of-Minds for\\ Human Simulation}

\author{%
  Pranav Dahiya \\
  Semilattice \\
  \texttt{p.dahiya@semilattice.ai} \\
}

\begin{document}

\maketitle

\begin{abstract}
  Predicting how a population will answer a new question is a long-standing goal.
  Statistical methods succeed at the level of the mass but falter at the level of
  the individual. Large language model simulators inherit this gap. They recover a
  population's central tendencies while flattening its heterogeneity, and they
  carry social biases and prompt brittleness that distort individual predictions.
  This paper introduces Anacreon, an audience simulation model that targets the
  individual level within a narrow, well-specified domain. Anacreon learns an
  authorship embedding that separates individuals, clusters a real qualitative
  corpus around seed people, and trains a dedicated adapter for each cluster, a
  mixture of minds, on a Gemma~4 12B base. It harvests demographics, psychological traits, and survey
  responses from public text, and augments each record with a chain-of-emotion. It
  reduces prompt brittleness by shuffling response options and reduces positive
  bias by balancing the training distribution. On a large, externally sourced
  survey, Anacreon reaches a state-of-the-art ordinal alignment of $0.775$, the
  individual-level accuracy measure on which the field has converged, with a small
  residual bias. The work is a step toward drawing aggregate insight from
  faithfully simulated individuals.

  {\itshape\small
  \begin{flushright}
  ``Dasein is a being that does not simply occur among other beings. Rather it
  is ontically distinguished by the fact that in its being this being is
  concerned about its very being.''\\[2pt]
  {\upshape---Heidegger\footnote{From \emph{Being and Time}, trans.\ Joan
  Stambaugh, revised by Dennis J.\ Schmidt (Albany: SUNY Press, 2010).}}
  \end{flushright}}
  \vspace{0.3em}
\end{abstract}

\section{Introduction}
\label{sec:introduction}

\begin{figure}[!ht]
	\centering
	\includegraphics[width=\linewidth]{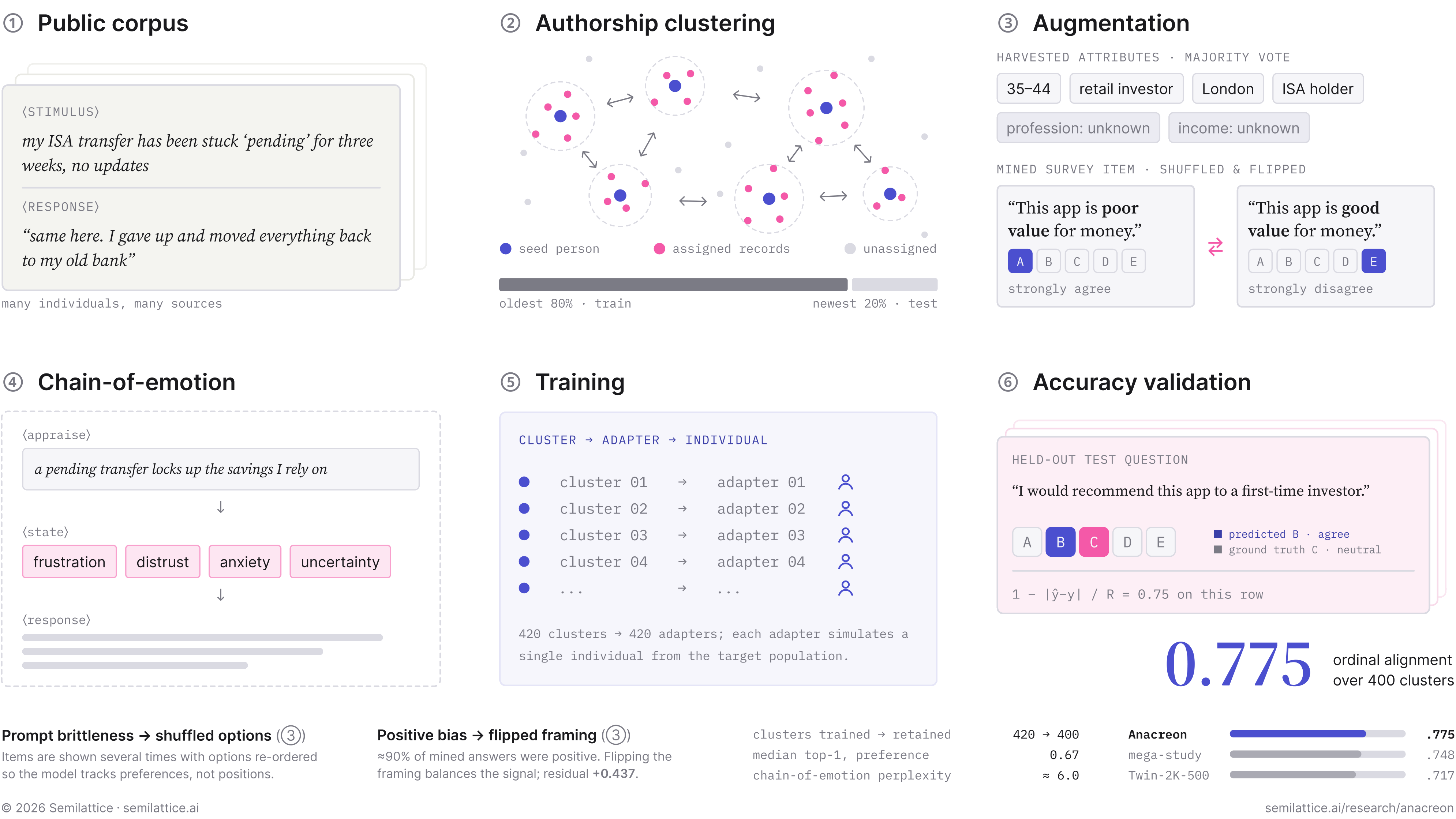}
	\caption{Overview of the Anacreon pipeline. Public text is clustered by author
		into a mixture of per-cluster models. Each cluster's records are augmented with
		harvested attributes and a chain-of-emotion, and a dedicated model is
		fine-tuned and selected against survey accuracy, reaching a state-of-the-art
		ordinal alignment of $0.775$.}
	\label{fig:pipeline}
\end{figure}

The ambition to render human behaviour calculable begins with death.
In 1662, John Graunt, now regarded as the father of demography,
distilled statistical regularity from London's weekly bills of mortality. His work
was the first empirical life table. Graunt's innovation was epistemological. He
saw that individual deaths are unpredictable while death in aggregate is
orderly~\citep{graunt1662}.
The theory became an industry in 1762 with the establishment of the 
Society for Equitable Assurances on Lives and Survivorships in
London~\citep{ogborn1962equitable}. The foundation laid by Graunt now underpins a vast global insurance
industry. Its worldwide premiums reached a record of about USD 7.1 trillion in
2023~\citep{swissre2024world}.

In 1835 Adolphe Quetelet proposed the idea of \emph{l'homme moyen}, the average
man about whom human measurements cluster~\citep{quetelet1835}. He imported the
astronomer's law of error, the normal distribution, into the study of human
beings. He found that many human traits stayed numerically stable from year to
year, among them height, weight, strength, and, more controversially, rates of
drunkenness. The statistics of demography reached its modern state in the first
half of the 20th century with the innovation of scientific opinion polling. Polling showed that a small,
well-drawn sample could out-predict a far larger but haphazard one. Leading up to the 1936 United States presidential election, the \emph{Literary
Digest} mailed out 10 million mock ballots and, from $2.4$ million
returns, forecast that Alfred Landon would defeat Franklin
Roosevelt~\citep{squire1988literary}. Roosevelt won in a landslide. The
magazine's vast sample was drawn from telephone and automobile registries, which
skewed affluent and Republican, so its size did not offset its bias. George
Gallup, using a quota sample of only 50{,}000, correctly
called Roosevelt's victory and even predicted the magnitude of the \emph{Digest}'s
error~\citep{gallup1940pulse}.

The 20th century did not stop at sampling. The Cold War defence-research
complex recast strategic human behaviour as formal
mathematics~\citep{vonneumann1944}. It also reimagined society as something that
could be simulated rather than merely sampled.  Founded in 1959, the Simulmatics Corporation built
``the People-Machine'' that sorted the electorate into voter
types and predicted their responses~\citep{lepore2020ifthen}. It advised the 1960 Kennedy campaign and then
claimed credit for the win. Its confidence eventually outran its evidence. \citet{lepore2020ifthen} shows that
many of its predictions were common sense dressed in the authority of the
machine. When the New York Times hired it to model the 1962 midterm elections,
the effort collapsed into a disorganised shambles. So did its contract with DARPA,
where it was hired to win over the South Vietnamese to the American cause. Its overreach prefigures a failure of modern LLM-based simulation that
\cref{sec:related_work} reviews. Simulmatics promised to model everything from
buying a dishwasher to countering an insurgency~\citep{lepore2020ifthen}, a
breadth so vast that its predictions could rarely be grounded or tested. 

Agent-based models (ABMs), brought to prominence with the Schelling
segregation model~\citep{schelling1971}, flipped the script by modelling
populations from the bottom up instead of the top down. Schelling demonstrated how
simple hand-set rules about individual preferences could lead to observed
collective patterns that no one intended.
Joshua Epstein and Robert Axtell generalised this insight
into artificial societies whose inequality, migration, and trade emerged from
local rules alone~\citep{epstein1996growing}. Such models show how aggregate
emergence can be simulated even with narrow insights into individual behaviour. 

Across this arc, aggregate trends have proved predictable only when they are
stable. A stable trend is not driven by any one person, so statistics captures it
well. Insurance risk and public-health rates stay orderly from year to year,
without any model of an individual. Transient trends resist this. Election results
and approval ratings turn on particular people at a particular moment, so they
cannot be deduced from stable regularities. They have to be measured, which is why
scientific opinion polling was developed. Agent-based modelling appears to bridge
the two, but it delivers less than it promises~\citep{windrum2007validation}. As
\citet{smaldino2017models} argues, an agent-based model shows only that a mechanism
is sufficient to produce an observed pattern. It explains how a pattern could have
arisen, not what an unobserved population would actually do. The
individual--aggregate gap therefore persists. It would close only if individual
behaviour could be predicted well enough across a representatively sampled
population, and this becomes tractable when the behaviour is restricted to a
narrow, well-specified domain.

The wider the target, the less any
prediction can be grounded against real data. Existing LLM simulators run
headlong into this limit. They repeat the overreach of ``the
People-Machine''~\citep{lepore2020ifthen} when they ask one model to stand in for
any population on any question. They recover a population's central tendencies while collapsing the heterogeneity,
uncertainty, and minority positions that distinguish real individuals within it.
This flattening is not incidental. To become useful assistants, these models
undergo extensive reinforcement learning from human feedback, which sharply reduces
the diversity of their outputs~\citep{kirk2024rlhf}. A better assistant is a worse
human simulator, since it answers as one agreeable average rather than as the many
distinct people a population contains. The base LLM also carries measurable social
biases~\citep{santurkar2023whose, gallegos2024bias}, and these systems rarely
correct them. \textit{Anacreon's goal is to simulate independent, heterogeneous
individuals well enough to draw meaningful aggregate insights.}  It simulates narrow, well-specified domains, and it
works to remove the model's bias rather than trust it. A machine
trusted to know more than it does is more dangerous than one that admits its
limits.

People reveal a great deal about themselves through the records they
leave in public. From a dataset of more
than 58{,}000 volunteers, \citet{kosinski2013private} showed that Facebook ``likes'' alone predict
many private attributes. The likes recover a person's Big Five (OCEAN) personality
traits, and they distinguish Democrats from Republicans in $85\%$ of cases.
Public behaviour is therefore a strong signal for private disposition. Anacreon relies on this signal. It collects what a population makes public and
infers the private traits that shape its answers. Rather than a single dense
model, it trains a \emph{mixture of minds}. It
addresses two failure modes that the literature has repeatedly documented in
naive LLM simulation (\cref{sec:related_work}). The first is \emph{prompt
brittleness}: simulated responses shift with the exact wording, framing, and
option ordering of the prompt rather than tracking stable preferences. The
second is \emph{bias}: models over-represent dominant groups and caricature
others through demographic stereotyping. On a large, held-out, survey test set, Anacreon achieves state-of-the-art ordinal
alignment.
\section{Background}
\label{sec:related_work}

This section traces how LLM simulators moved
from matching populations to modelling individuals. It outlines why faithful individual
prediction remains hard, how bias distorts the groups being simulated, and how
the field has learned to measure what remains.

\subsection{From aggregate to individual fidelity}
\label{sec:rw_populations}

The earliest LLM simulators targeted populations, not persons. Prompted models
could reproduce classic findings in aggregate. \citet{argyle2023silicon} coined
the term \emph{silicon samples}. They conditioned a model on real
socio-demographic backstories and recovered the opinion distributions of many
subgroups, a property they named \emph{algorithmic fidelity}. Their evaluation
was explicit about its scope. It matched the aggregate distribution but did not
assess individual-level correspondence. Population-level prototyping followed the
same logic, expanding a few seed personas into a community to surface plausible
interactions \citep{park2022social}.

Attention then turned to identifiable individuals. \emph{Generative Agents}
gave the template, producing coherent longitudinal behaviour for sandbox
characters \citep{park2023generative}. \citet{park2024thousand} scaled this to
1{,}052 real people, grounding an agent for each in a two-hour interview and
structured surveys. Their agents recovered roughly $85\%$ of a participant's own
two-week test--retest consistency, and richer self-report data narrowed accuracy
gaps across subgroups relative to demographic prompting, which is known to induce
stereotyping \citep{santurkar2023whose}.

Rich grounding, however, buys aggregate plausibility more cheaply than individual
heterogeneity. The \emph{Twin-2K-500} twins reached a $0.717$ individual-level
accuracy score, $87.7\%$ of the human ceiling, yet expressed markedly less response
variance than humans \citep{twin2k2025}. A large follow-up study on the same
individuals was more sobering. On novel stimuli, twin responses correlated only
weakly with their human counterparts (average $r \approx 0.20$) and fell back
toward the population mean \citep{peng2025megastudy}. They compress variation, and a large share of regression
coefficients estimated on them differ in size or even flip in sign
\citep{bisbee2024synthetic}. A good aggregate fit does not protect the
individual-level relationships an analysis depends on.

These distortions motivate a shift from prompting frozen models to training
grounded ones. Finetuning on large corpora of human responses improves
distributional alignment, though it exposes a tension. Better distribution
matching can reduce single-response accuracy \citep{kolluri2025finetuning}. The
same approach may scale to a single trained model of behaviour across many
experiments \citep{binz2024centaur}.

\subsection{Bias and group representation}
\label{sec:rw_bias}

A simulated population can misrepresent the very groups it stands in for.
\citet{wang2025flatten} identify two failure modes. The first is
\emph{misportrayal}. Prompted with a demographic identity, a model more often
reflects what outsiders think of a group than what its members say about
themselves, and this falls hardest on marginalised identities. The second is
\emph{flattening}. The model renders each group as homogeneous and ignores
within-group heterogeneity. These effects compound the demographic misalignment
of \citet{santurkar2023whose} and the variance compression noted above.

\citet{hu2025population} treat bias as a property to be corrected before
simulation. They generate narrative personas from human corpora and reweight them
by importance sampling so that the persona set's Big Five distribution matches a
global human reference. This lowers the distributional distance to that reference
by about $32\%$ relative to the strongest prior persona set. However, this still fails 
to address the failure modes highlighted above.

\subsection{Measuring the gap}
\label{sec:rw_metric}

Comparing simulators requires a alignment metric for the ordinal and Likert scales that are
pervasive in surveys, where binary exact-match is uninformative. 
\citet{twin2k2025} and \citet{kolluri2025finetuning} adopt what this paper terms
\emph{ordinal alignment}. Given a predicted response $\hat{y}$ and a ground truth
$y$ on a scale of range $R$ (for an $n$-point scale, $R = n-1$), accuracy is one
minus the normalised mean absolute deviation, averaged over questions,
\begin{equation}
\label{eq:ordinal_alignment}
\mathrm{A_{ord}}
= 1 - \frac{\mathrm{MAD}}{R}
= 1 - \frac{1}{N}\sum_{i=1}^{N}
\frac{\lvert \hat{y}_i - y_i \rvert}{R}.
\end{equation}
The measure lies in $[0,1]$. It equals $1$ for an exact match and $0$ when the
prediction is maximally wrong, and it reduces to ordinary accuracy on binary
items. For example, on a $1$-to-$7$ item, a predicted $6$ against a true $5$
scores $0.833$. It has become the community's settled measure of individual-level
fidelity, adopted by the Twin-2K-500 dataset, its mega-study follow-up, and the
normalised-deviation accuracy of \citet{kolluri2025finetuning}. On this common
scale, the state of the art still sits well short of human reliability
\citep{park2024thousand, twin2k2025}.

Ordinal alignment is necessary but not sufficient. A model can score well on
\cref{eq:ordinal_alignment} while collapsing the population's variance, so the
field often pairs it with distribution-level checks on the spread and shape of
predicted responses. 
\section{Method}
\label{sec:method}

Anacreon is trained on unstructured text drawn from many sources. Each record is
cast as a \emph{(stimulus, response)} pair, where the stimulus, $s$, is a condition
and the response, $r$, is the outcome an individual produced under it. The stimulus
space is defined by the target population to be modelled. Rather than map the
stimulus straight to the response, Anacreon first constructs a
\emph{chain-of-emotion} ($c$), a short trace of the emotional states and appraisals
the stimulus evokes, then produces the response conditioned on that trace
(\cref{sec:chain_of_emotion}). Fine-tuning minimises the expected negative log-likelihood of the
target,
\begin{equation}
  \mathcal{L}(\theta)
  = -\,\mathbb{E}_{(s, c, r) \sim \mathcal{D}}
      \big[\log p_{\theta}(c, r \mid s)\big],
  \qquad
  p_{\theta}(c, r \mid s) = p_{\theta}(c \mid s)\,p_{\theta}(r \mid s, c),
  \label{eq:cot}
\end{equation}
where $\mathcal{D}$ is the training set and $\theta$ the model parameters.

\subsection{Clustering}
\label{sec:seed}
\label{sec:clustering}

A subset of individuals with rich data is chosen as a seed population. Inclusion
requires a minimum quantity of data per individual. 
A transformer is then trained to embed individuals so that the seed people
are maximally separated in the embedding space. Training uses the contrastive
authorship-representation objective of \citet{rivera2021luar}. Each individual in
a batch contributes two disjoint samples of their text, which the encoder maps to
embeddings $u_i$ and $v_i$. The objective pulls the two samples of the same
individual together and pushes samples of different individuals apart. For a batch
of $B$ individuals, with cosine similarity
$s(x, y) = x^{\top} y / (\lVert x \rVert \, \lVert y \rVert)$ and temperature
$\tau$, the loss is the symmetric contrastive (InfoNCE) term
\begin{equation}
  \mathcal{L} = -\frac{1}{2B} \sum_{i=1}^{B} \left[
    \log \frac{\exp\!\big(s(u_i, v_i) / \tau\big)}
              {\sum_{j=1}^{B} \exp\!\big(s(u_i, v_j) / \tau\big)}
    + \log \frac{\exp\!\big(s(u_i, v_i) / \tau\big)}
                {\sum_{j=1}^{B} \exp\!\big(s(u_j, v_i) / \tau\big)}
  \right].
  \label{eq:contrastive}
\end{equation}
The other individuals in the batch act as negatives. Minimising \cref{eq:contrastive}
separates individuals in the embedding space, which is what the clustering step
below relies on.

The trained model is then applied to the full corpus. Every record is assigned to its
nearest seed individual, which partitions the corpus into clusters. Each cluster
is then split by time. The newest $20\%$ of records is held out for testing, and
the remaining $80\%$ is used for training. The temporal split prevents later
information from leaking into training.

\subsection{Data Augmentation}
\label{sec:attributes}

\begin{figure}[htbp]
  \centering
  \includegraphics[width=\linewidth]{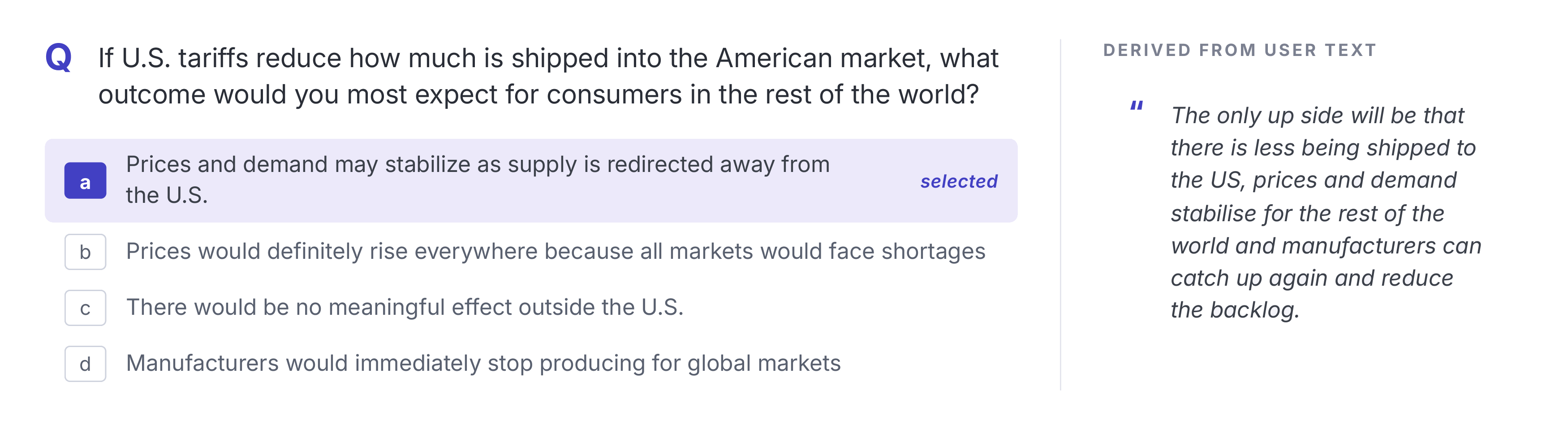}
  \caption{Generation of a mined survey question.}
  \label{fig:survey_example}
\end{figure}

An LLM-as-judge extracts attributes from each cluster's text. These include
demographics and a set of psychological traits. A proprietary harness
reduces hallucination, and a majority vote across multiple runs stabilises each
attribute. The harness favours accuracy over coverage. When the runs disagree too
much for a majority to form, the attribute is left undefined rather than guessed.
This is why several demographic categories in \cref{sec:results} carry an unknown
bucket. Survey questions are also mined from the text, since some text forms
reveal answers to standard instruments. \Cref{fig:survey_example} shows one
example, where user generated free-text is turned into a multiple-choice question whose
answer is known. Survey questions are mined separately on the train
and test splits, so that test answers cannot leak into training. The mined survey
set spans two formats: ordinal, Likert-style and
categorical multiple choice questions.

Due to the positivity bias in LLM responses~\citep{syceval2025}. most of the ordinal questions generated
by the mining agent had positive responses, skewing towards "agree"/"strongly agree"
in 90\% of cases. To fix this problem, every question with a non-neutral response was
included twice in the training and test sets, once framed positively and once negatively.
This not only aims to address the sycophancy problem but also judges how susceptible the
trained models will be to prompt brittleness.

\subsubsection{Chain-of-Emotion}
\label{sec:chain_of_emotion}

Each record is also augmented with a \emph{chain-of-emotion}, in the sense of the
appraisal-based construction of \citet{croissant2024chainofemotion} and the
latent emotional-state alignment of \citet{wu2025humanlm}. An agent harness takes the
stimulus and selects the harvested attributes relevant to it. Only a few
attributes matter for any given stimulus, not all. This is consistent with the
constructive-choice view, in which people weight a context-dependent subset of
attributes~\citep{bettman1998constructive}. The chain then maps the evolution of emotional state
from stimulus to response. It is constructed for both unstructured text and mined survey questions.

\subsection{Training and Checkpoint Selection}
\label{sec:training}

A separate high-rank LoRA adapter is trained for each cluster on its
augmented data, on top of a Gemma~4 12B base model~\citep{gemma4report},
using QLoRA~\citep{dettmers2023qlora} for compute efficiency. Together the
per-cluster adapters form Anacreon's mixture of minds, one model specialised to
each cluster. At
rank 64, the 12B base yields about 260M trainable parameters per
cluster. Summed over the 420 trained clusters, Anacreon learnt about 110B parameters in total. 
During each training epoch, the order of responses for each survey question in the training set is
randomly shuffled. This acts as a sort of dropout to ensure the model is not learning to predict responses
purely based on their position in the list of options.

\label{sec:cluster_quality}%
The clusters vary in quality, as expected. Coherent
clusters, whose members are genuinely similar, train well. Incoherent clusters
train poorly. In the worst case a cluster contains conflicting statements, so the
model cannot tell whether a response should be positive or negative. A long tail
of clusters failed to converge and was pruned.
At inference the 400 retained per-cluster models unpack to ~4.9T parameters.

The best checkpoint is chosen by balancing three quantities on held-out data. The
first is top-1 accuracy $A_{\mathrm{mc}}(\theta)$ on the multiple-choice questions,
the true positive rate. The second is the ordinal alignment
$A_{\mathrm{ord}}(\theta)$ of \cref{eq:ordinal_alignment} on the ordinal questions.
The third is the perplexity of the chain-of-emotion. For the target tokens
$w_{1:N}$ of a held-out example, only the chain-of-emotion,
perplexity is the exponentiated mean per-token negative log-likelihood,
\begin{equation}
  \mathrm{PPL}(w_{1:N}) = \exp\!\left(-\frac{1}{N} \sum_{i=1}^{N}
    \log p_{\theta}(w_i \mid w_{<i}, s)\right),
  \label{eq:perplexity}
\end{equation}
the exponentiated cross-entropy between the model and the
target~\citep{jelinek1977perplexity}. A low perplexity means the model has
converged well on next-token prediction of the chain-of-emotion. Writing
$\mathrm{PPL}(\theta)$ for its mean over the held-out pairs, the three signals are
combined into a selection score, and the checkpoint that maximises it is kept,
\begin{equation}
  \theta^{\star} = \operatorname*{arg\,max}_{\theta}\;
    \alpha \, A_{\mathrm{mc}}(\theta) + \beta \, A_{\mathrm{ord}}(\theta)
    - \gamma \, \mathrm{PPL}(\theta),
  \label{eq:checkpoint}
\end{equation}
where the nonnegative weights $\alpha, \beta$ and $\gamma$ balance accuracy, ordinal alignment,
and convergence.
\begin{figure}[b]
	\centering
	\begin{subfigure}[t]{0.49\linewidth}
		\centering
		\includegraphics[width=\linewidth]{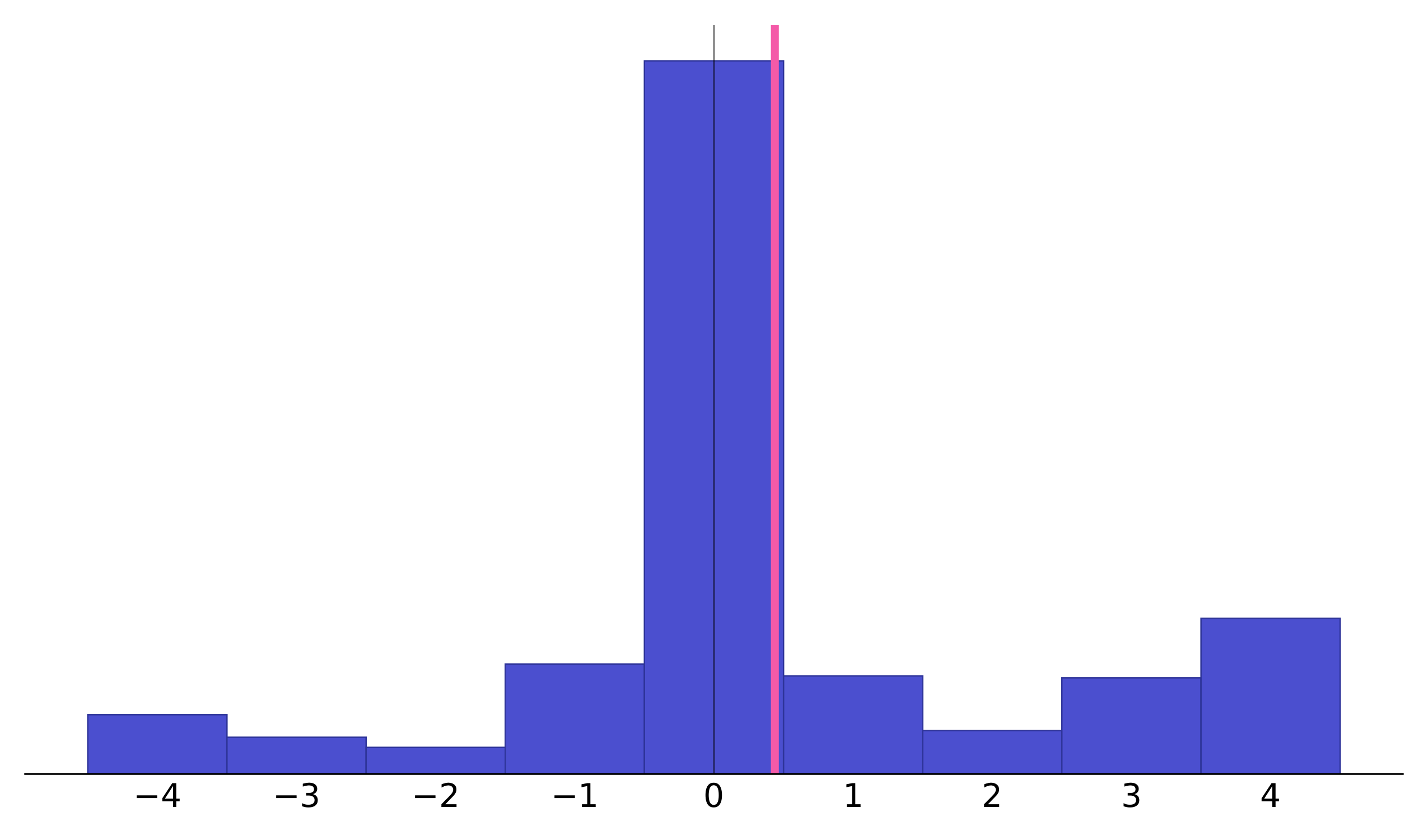}
		\caption{Signed error, pooled over all rows.}
		\label{fig:signed_pooled}
	\end{subfigure}
	\hfill
	\begin{subfigure}[t]{0.49\linewidth}
		\centering
		\includegraphics[width=\linewidth]{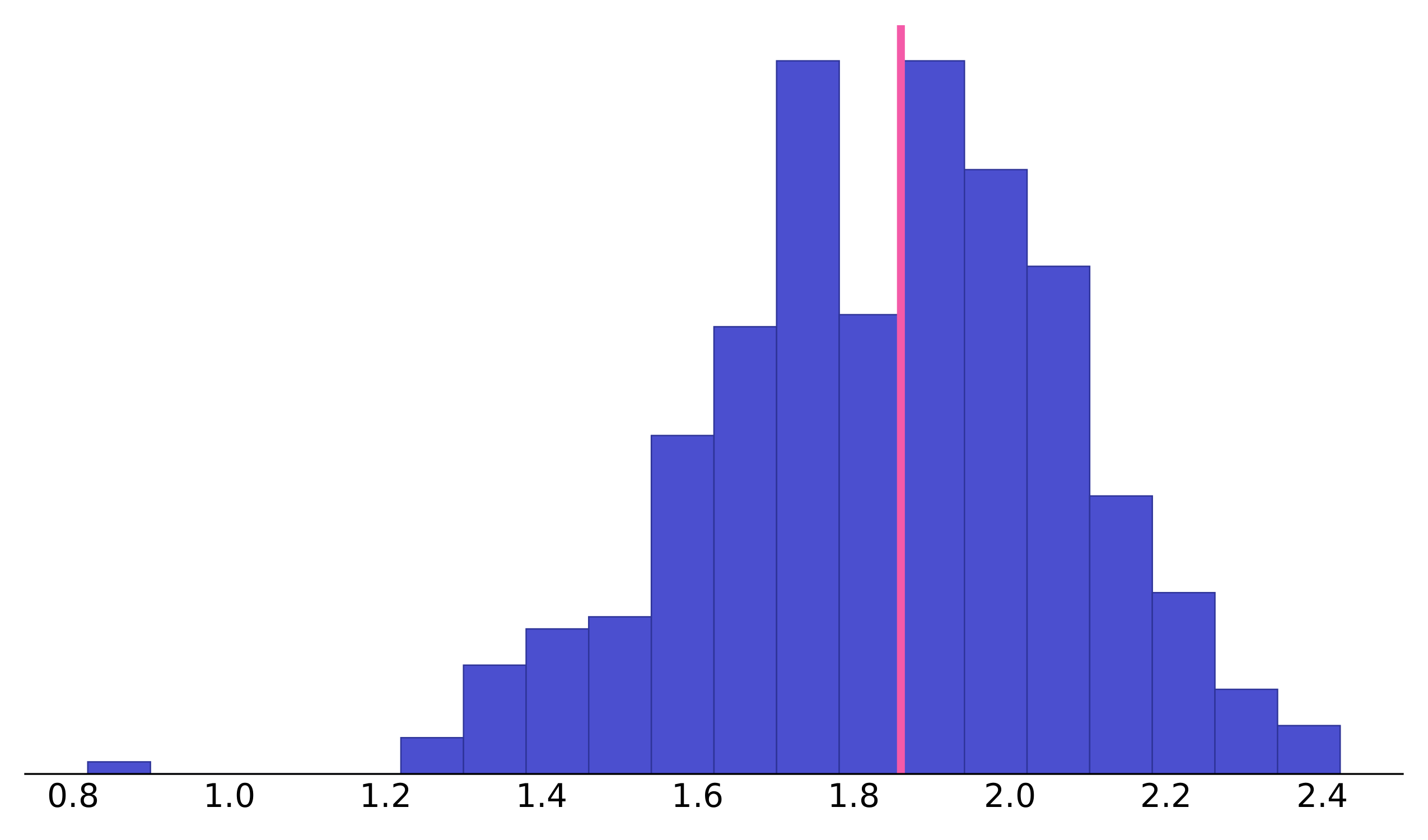}
		\caption{Per-cluster standard deviation $\sigma$.}
		\label{fig:signed_std}
	\end{subfigure}
	\caption{Signed prediction error, in semantic scale positions. (a) pooled over
		all survey rows, peaked at zero with mean $+0.437$. (b) per-cluster standard
		deviation $\sigma$ of the signed error, with a median near $1.85$.}
	\label{fig:signed}
\end{figure}

\section{Results}
\label{sec:results}

This section evaluates Anacreon on a B2B audience model built for
a Semilattice customer. The modelled individuals are SME merchants who take
physical, in-person card payments.

\subsection{Bias}
\label{sec:setup}
\label{sec:bias}
\label{sec:analysis}

A well-calibrated simulator should place its predictions symmetrically around the
true responses. Base LLMs instead tend toward a positive, sycophantic
bias~\citep{syceval2025}, which Anacreon is designed to reduce. \Cref{fig:signed}
reports the signed prediction error, measured in semantic scale positions.
\Cref{fig:signed_pooled} pools all survey rows. Each model was prompted with each
row in its test survey set 5 times with shuffled responses. The distribution concentrates at
zero but retains a small positive mean of $+0.437$. A positivity bias therefore
remains, but it is much reduced relative to a base LLM. 

\begin{figure}[t]
	\centering
	\begin{subfigure}[t]{0.49\linewidth}
		\centering
		\includegraphics[width=\linewidth]{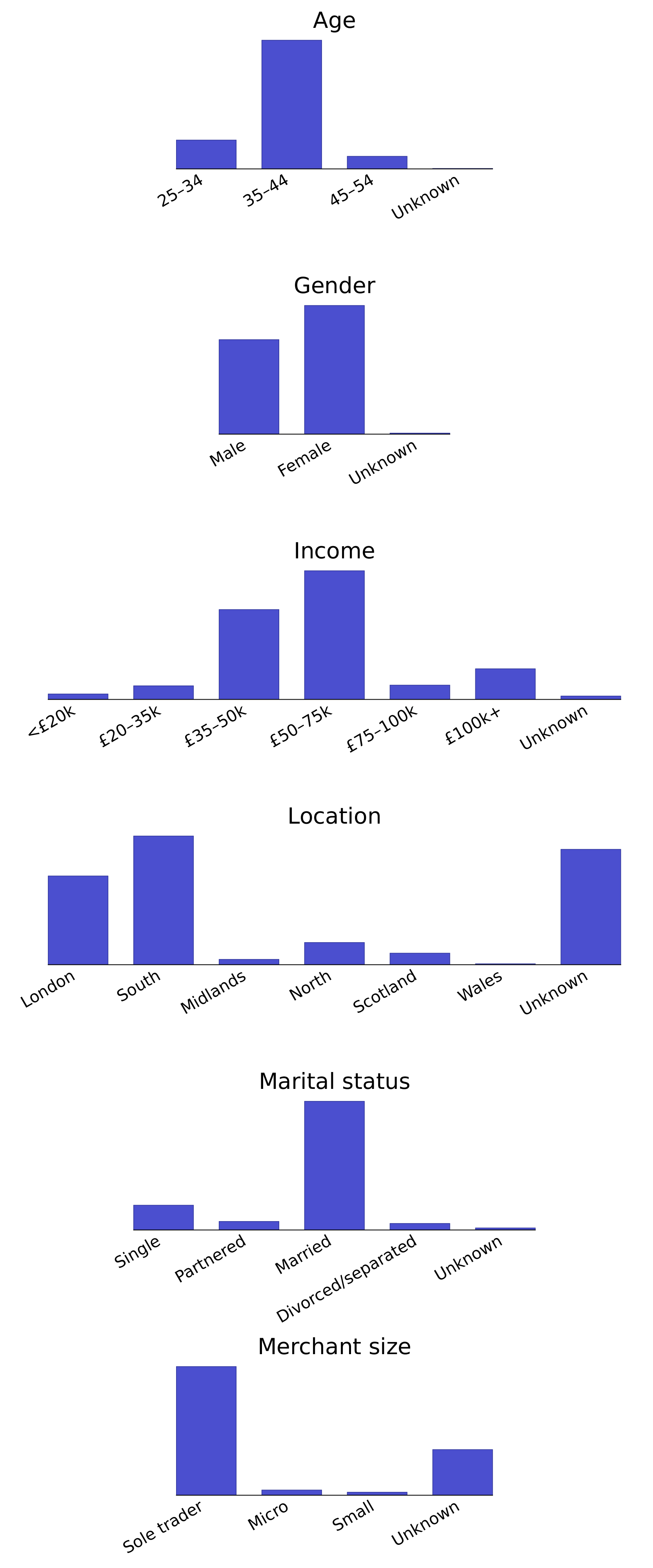}
		\caption{Full modelled population.}
		\label{fig:demographics_full}
	\end{subfigure}
	\hfill
	\begin{subfigure}[t]{0.49\linewidth}
		\centering
		\includegraphics[width=\linewidth]{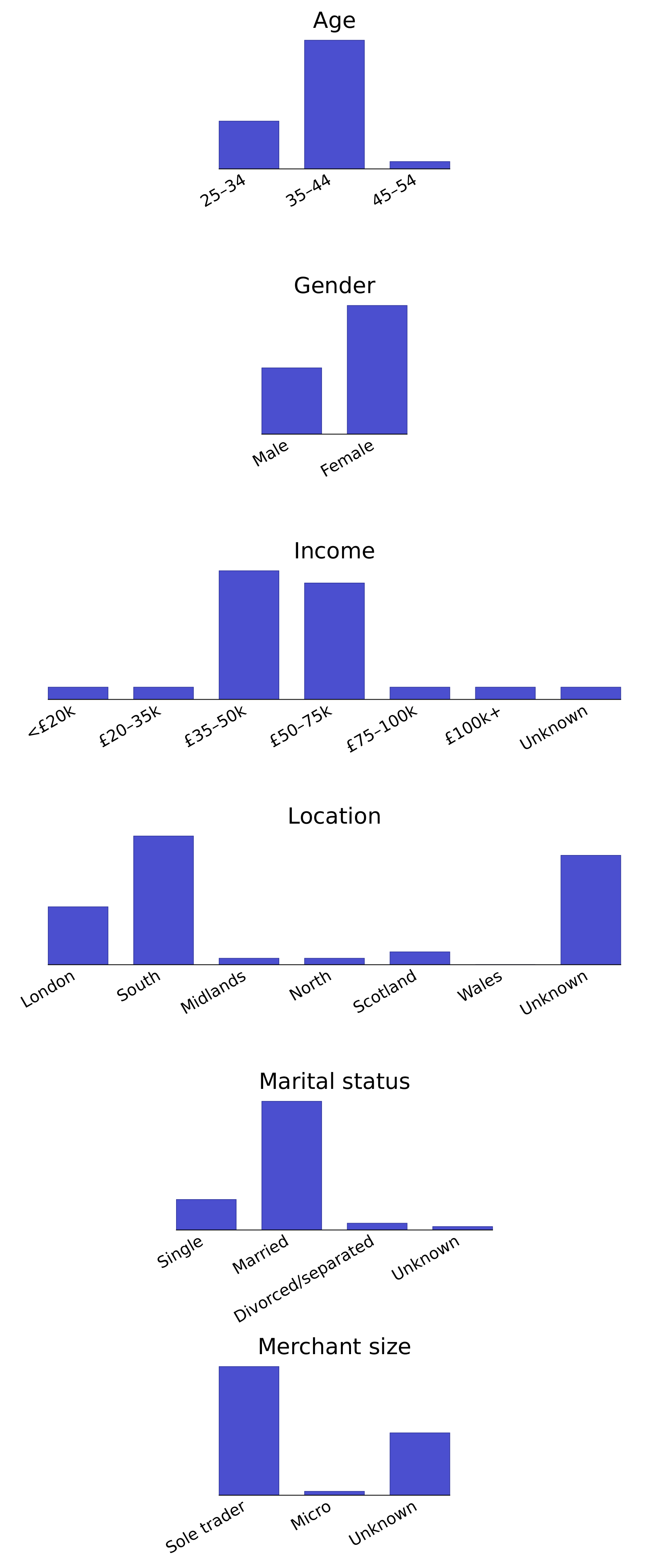}
		\caption{The 50 lowest-scoring clusters}
		\label{fig:demographics_bottom50}
	\end{subfigure}
	\caption{Demographic composition over age, gender, income, location, marital
		status, and merchant size. (a) the full modelled population; (b) the 50
		clusters with the lowest ordinal alignment. The two distributions are
		similar, so the weakest clusters are not concentrated in any single
		demographic group.}
	\label{fig:demographics}
\end{figure}

The errors vary in spread. \Cref{fig:signed_std} shows the standard
deviation of the signed error within each cluster. The distribution is unimodal,
with a median near $1.85$ scale positions. This spread, rather than the small
systematic bias, accounts for most of a cluster's deviation from the true
responses.

\Cref{fig:demographics_full} shows the demographic composition of the modelled
population across age, gender, income, location, marital status, and merchant
size. The spread is broad and is not dominated by any single group. This
indicates that the attribute-mining method (\cref{sec:attributes}) did not
introduce a strong sampling bias.

\Cref{fig:demographics_bottom50} shows the same composition for the 50
lowest-scoring clusters, ranked by ordinal alignment. Its shape tracks the full
population (\cref{fig:demographics_full}). The weakest clusters are therefore not
concentrated in any single demographic group. This suggests that poor performance
reflects the data quality within a cluster rather than the demographic it
represents.

\subsection{Survey Accuracy}
\label{sec:main_results}

\begin{figure}[htbp]
	\centering
	\begin{subfigure}[t]{0.49\linewidth}
		\centering
		\includegraphics[width=\linewidth]{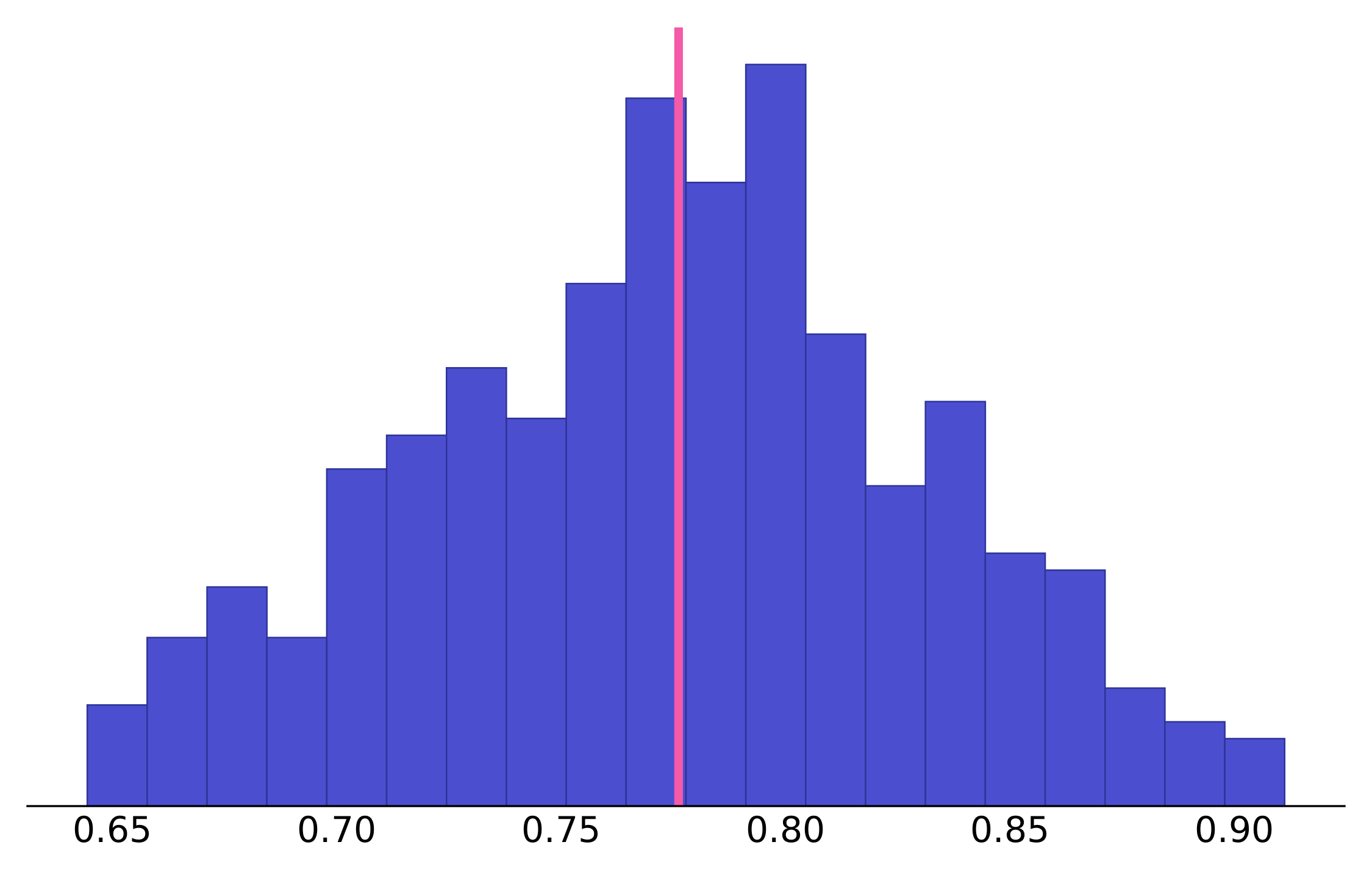}
		\caption{Ordinal alignment distribution}
		\label{fig:ordinal}
	\end{subfigure}
	\hfill
	\begin{subfigure}[t]{0.49\linewidth}
		\centering
		\includegraphics[width=\linewidth]{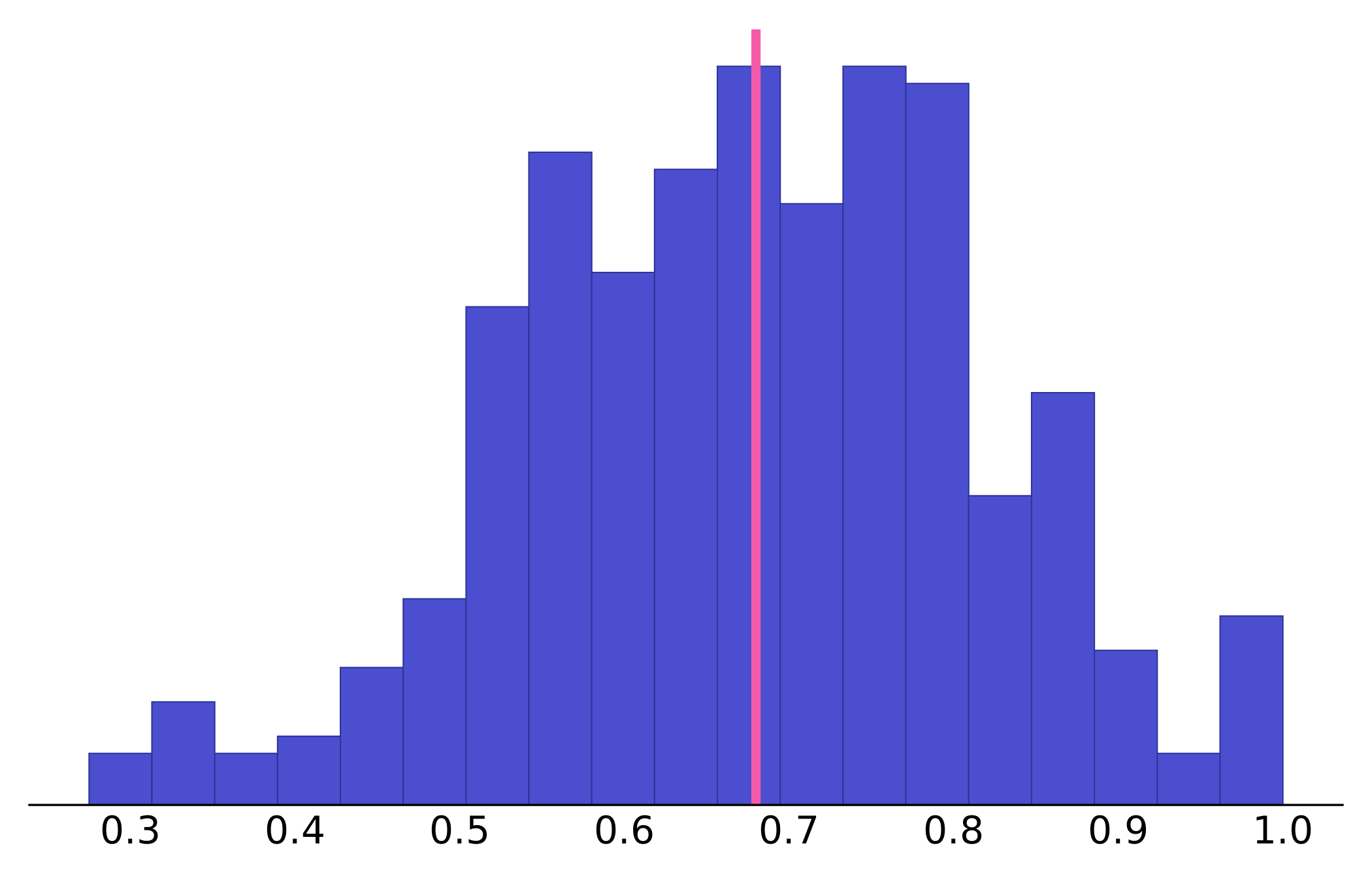}
		\caption{Top-1 accuracy distribution}
		\label{fig:mc_accuracy}
	\end{subfigure}
	\caption{Frequency distribution of per-cluster survey performance. (a) ordinal alignment ($1 - \mathrm{MAD}/R$) on the ordinal
		questions; the median cluster (pink line) scores $0.775$ (b) top-1 accuracy on the multiple-choice
		preference questions; the median cluster reaches $0.679$}
	\label{fig:survey}
\end{figure}

The mined survey set spans two formats, ordinal Likert-style items and
multiple-choice preference questions (\cref{sec:attributes}). \Cref{fig:survey}
reports both. \Cref{fig:ordinal} shows ordinal alignment across the evaluation
clusters. The distribution is centred at $0.775$, and its bulk sits above the
shuffle baseline band. \Cref{tab:comparison} places this score against prior work.
On the shared ordinal-alignment measure ($1 - \mathrm{MAD}/R$), Anacreon at
$0.775$ exceeds every prior system that reports it, namely Twin-2K-500 at $0.717$,
Socrates-Qwen-14B at $0.740$, and the digital-twin mega-study at $0.748$. This is
the state of the art on this measure.

\Cref{fig:mc_accuracy} reports top-1 accuracy on the multiple-choice preference
questions, that is, whether the predicted choice is exactly correct. The median
cluster reaches $0.679$. This is comparable to the $0.657$ exact-match accuracy of
the interview-grounded agents of \citet{park2024thousand}, the closest prior system
that reports an exact-match measure (\cref{tab:comparison}). The digital-twin
literature otherwise reports only ordinal questions.

\begin{table}[htbp]
  \centering
  \caption{Individual-level accuracy of Anacreon and prior work, grouped by the
    measure used; higher is better. The scores come from related but not identical
    measures on different datasets, so they are indicative rather than a controlled
    comparison. Top-1 (exact-match) accuracy compares Anacreon with the
    interview-grounded agents of \citet{park2024thousand}. Ordinal alignment
    ($1 - \mathrm{MAD}/R$, \cref{eq:ordinal_alignment}) compares it with the
    digital-twin systems that report it. The \emph{Approach} column summarises each
    system's modelling technique. The final column gives the human two-week
    test--retest (self-consistency) baseline where the paper reports one.}
  \label{tab:comparison}
  \vspace{6pt}
  \small
  \begin{tabular}{lllcc}
    \toprule
    Metric & Method & Approach & Score & Test--retest \\
    \midrule
    \multirow{2}{*}{Top-1 accuracy}
      & \citet{park2024thousand}       & interview-based prompts & $0.657$          & $0.795$ \\
      & \textbf{Anacreon}              & per-cluster fine-tuning & $\mathbf{0.679}$ & --      \\
    \midrule
    \multirow{4}{*}{Ordinal alignment}
      & \citet{twin2k2025}             & context-injected twins  & $0.717$          & $0.817$ \\
      & \citet{kolluri2025finetuning}  & supervised fine-tuning  & $0.740$          & --      \\
      & \citet{peng2025megastudy}      & twins, novel stimuli    & $0.748$          & --      \\
      & \textbf{Anacreon}              & per-cluster fine-tuning & $\mathbf{0.775}$ & --      \\
    \bottomrule
  \end{tabular}
\end{table}

\subsection{Chain-of-Emotion Perplexity}
\label{sec:cot_perplexity}

Anacreon produces a chain-of-emotion before each answer
(\cref{sec:chain_of_emotion}). Its perplexity (\cref{eq:perplexity}) measures how
well the model predicts these sequences. A value of $k$ means the model is, on
average, as uncertain as a uniform choice among $k$ tokens. A value that's too low indicates the model has become more deterministic than desired; a value that's too high indicates the model has failed to converge well.

Perplexity has no universal scale. It depends on the tokenizer, the vocabulary,
and the domain, so values compare only within a fixed setup. As a rough guide,
strong open-domain language models report word-level perplexities in the high
teens on standard English benchmarks, for instance $17.5$ on
WikiText-103~\citep{radford2019gpt2}. Single-digit values usually signal a narrow
or highly predictable domain.

\begin{figure}[htbp]
  \centering
  \includegraphics[width=0.5\linewidth]{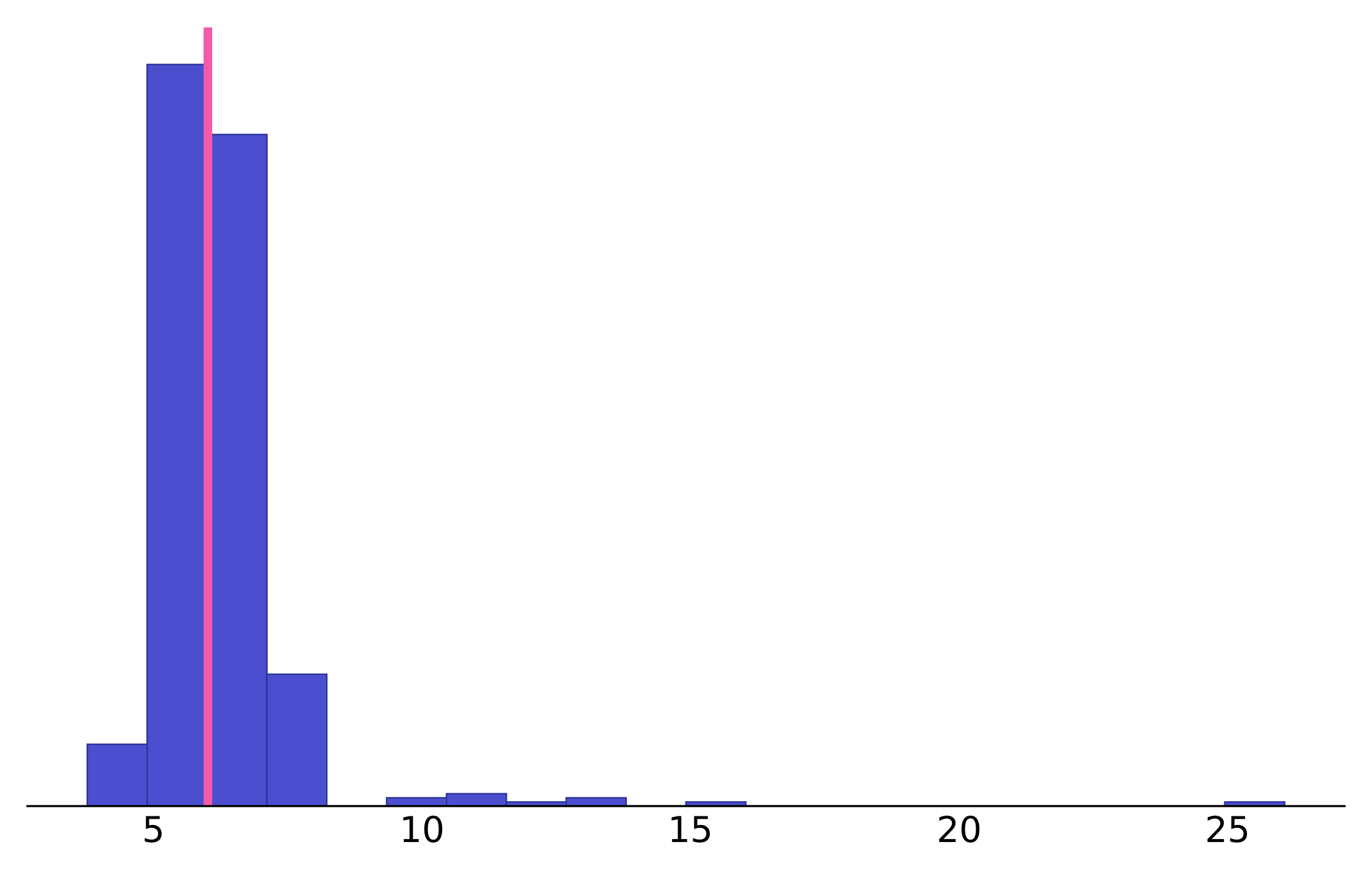}
  \caption{Chain-of-emotion perplexity frequency distribution per cluster. The
    vertical line marks the median, near $6.0$.}
  \label{fig:cot_perplexity}
\end{figure}

The chain-of-emotion is such a narrow domain. It is not free text. At each step the
continuation is drawn from a narrow, structured space of emotional states and
appraisals (\cref{sec:chain_of_emotion}), so few tokens are ever plausible. This
makes perplexity a sensitive convergence signal. When the valid continuations are
few, a converged model concentrates almost all of its mass on them and reaches a
low perplexity near the entropy of that space. A high perplexity would show the
opposite, that the model still spreads mass over tokens the format does not
permit. \Cref{fig:cot_perplexity} reports this quantity per cluster at the best
epoch. The distribution is unimodal and concentrated, with a median near $6.0$.
The models mostly sit in the single digits, well inside the range that such a
constrained target allows. This indicates that they reliably reconstruct the
emotional chain that leads to each answer.

\section{Discussion}
\label{sec:discussion}

\paragraph{Narrow Domain.}
Anacreon specifically targets narrowly defined populations, addressing the overreach that
\cref{sec:introduction} traced from Simulmatics to today's simulators. 
 The beliefs encoded in each cluster's training set
therefore act like a belief network~\citep{pearl1988}. That network is well
supported near the training domain and sparse away from it. Conclusions extend to
topics adjacent to the training data, where the network still carries evidence.
They cannot be trusted far beyond it, where the evidence thins to guesswork. At inference a query can be checked for
support in the training data, so an answer is offered only when the records provide some
basis for it. Anacreon can therefore recognise the edge of its own competence and
decline to speak past it. 

\begin{figure}[htbp]
  \centering
  \begin{subfigure}[t]{0.49\linewidth}
    \centering
    \includegraphics[width=\linewidth]{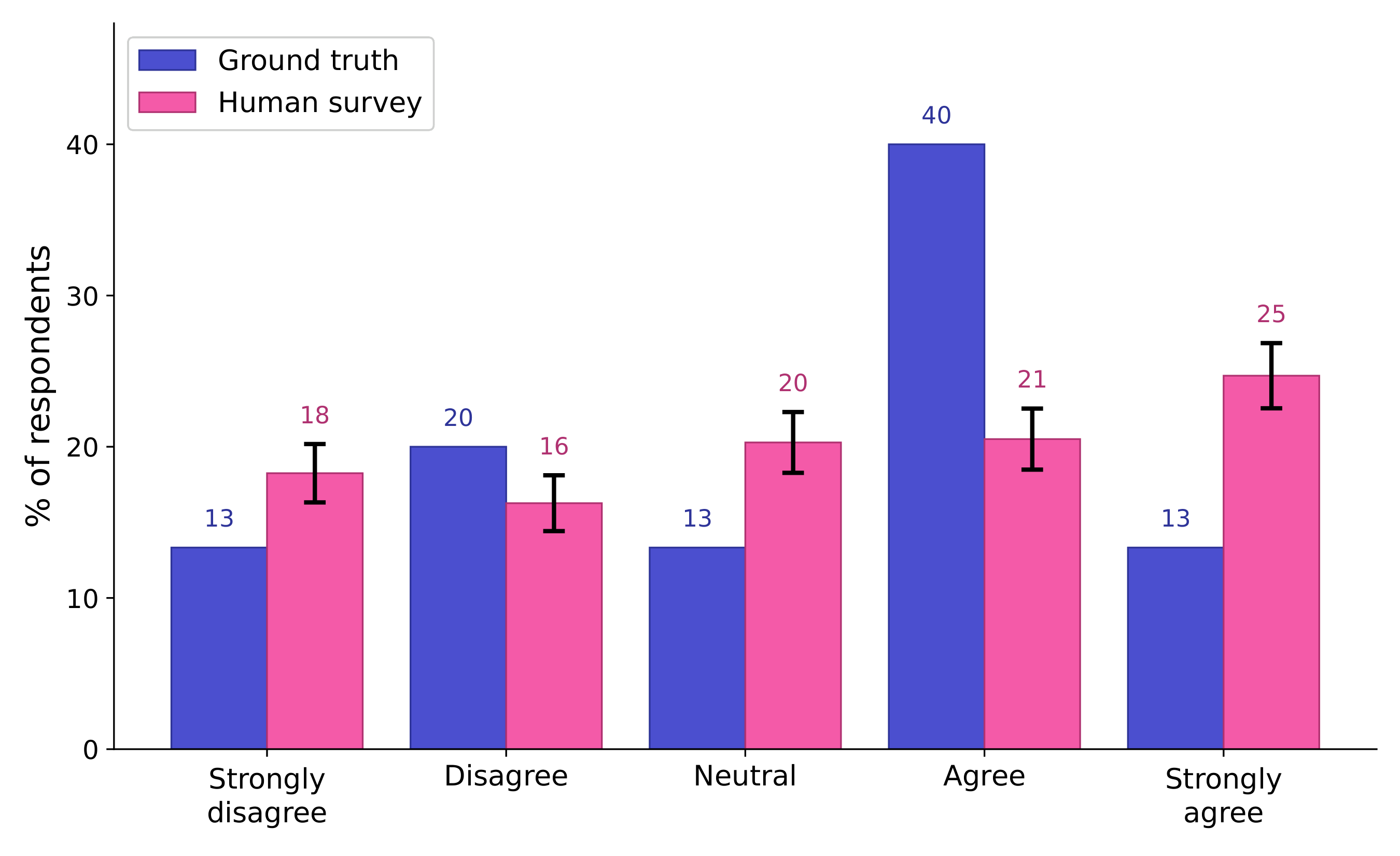}
    \caption{Human test-retest}
    \label{fig:likert_human}
  \end{subfigure}
  \hfill
  \begin{subfigure}[t]{0.49\linewidth}
    \centering
    \includegraphics[width=\linewidth]{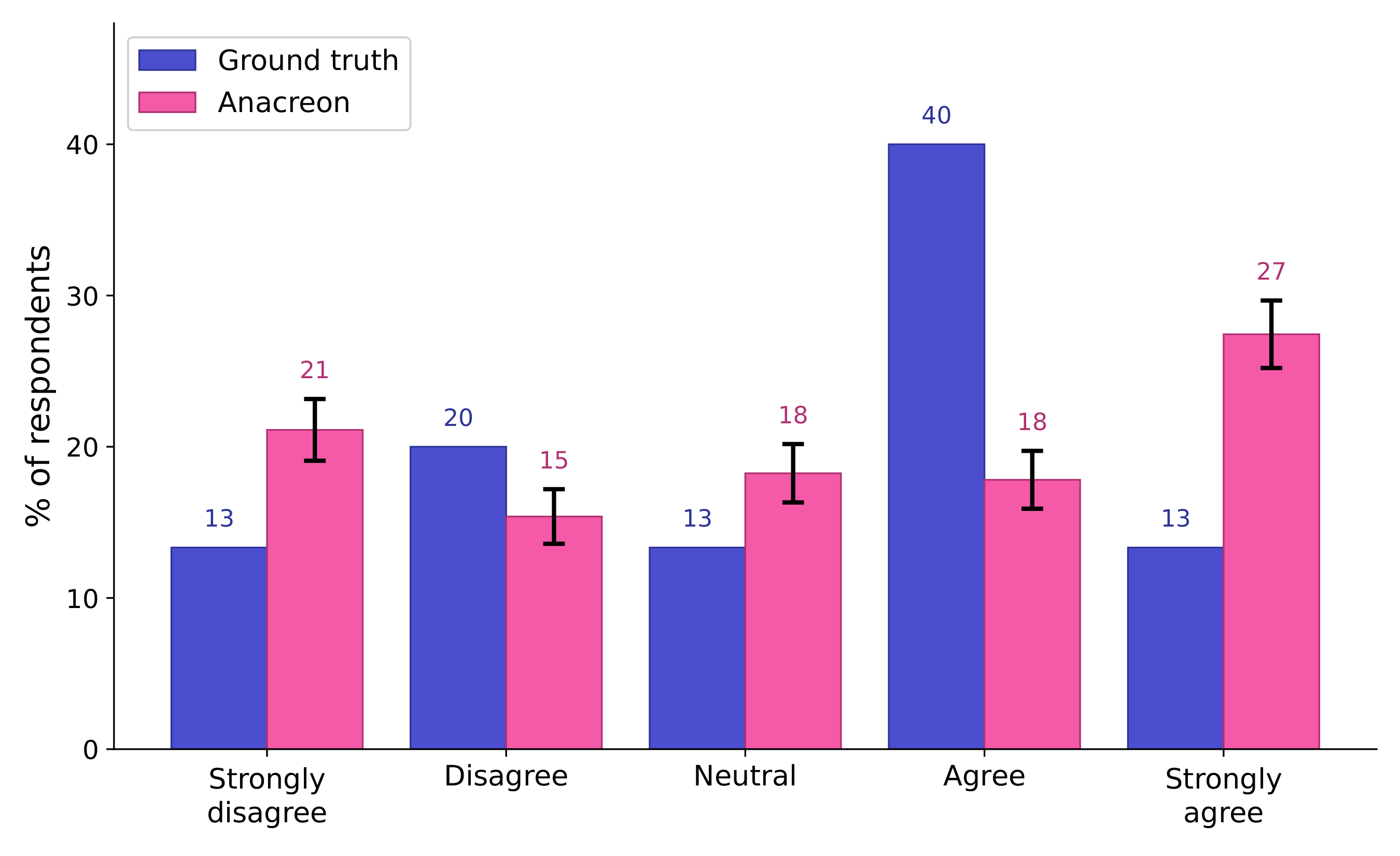}
    \caption{Anacreon}
    \label{fig:likert_model}
  \end{subfigure}
  \caption{Illustrative response distributions on a five-point Likert item, each
    against the same ground truth. The blue bars indicate a theoretical ground truth 
    response distribution that's being measured with (a) a human survey, and (b) an Anacreon simulation.
    Both charts indicate a worst case skew that could arise from the natural variance in human 
    responses and the measured median variance of Anacreon. Anacreon's skew is only moderately worse than human test-retest.}
  \label{fig:likert}
\end{figure}

\paragraph{Bias and Variance.}
A positivity bias, though small, has not been completely eliminated
(\cref{sec:bias}). More importantly, the evaluation so far concerns
individual-level accuracy, and despite state-of-the-art ordinal alignment and
top-1 accuracy, the models display high variance (\cref{fig:signed_std}). Humans
are not deterministic either. Single survey items measure attitudes with low
reliability, around $0.5$; re-surveyed respondents can give substantially different
answers~\citep{alwin1991reliability, achen1975mass}. By classical test theory, a
reliability of $0.5$ corresponds to a test-retest standard deviation near $1.27$
scale positions on a five-point item~\citep{lordnovick1968}. \Cref{fig:likert}
sets this human baseline against Anacreon. Anacreon's per-cluster median standard
deviation is about $1.85$, above the human level but only moderately so. A faithful
simulator should reproduce this human spread rather than collapse to a point, so
the target is the human level, not zero variance. However, while the median model converges
well, models at the tail end are only marginally better than a random guesser. 
Since the positivity bias was mostly eliminated, this should not stop Anacreon from being useful for
aggregate, population-level predictions.

\paragraph{From individuals to aggregates.}
The value of a faithful individual simulator lies in what it enables at the
aggregate. A population's spread, its disagreement, and
its tails often carry the signal that matters. The next step is to study how better
individual simulation translates into better aggregate prediction, by composing
many well-simulated individuals into an estimate of a population's distribution.
The template for doing so can come from Opinion Polling. A poll rarely
draws a perfectly representative sample. Survey researchers correct the imbalance
after the fact. They weight the respondents by post-stratification, and by its
multilevel-regression extension, so that the reweighted sample matches known
population margins~\citep{park2004mrp}. The same correction can be applied to
Anacreon's mixture of minds, reweighting the per-cluster models so that their
pooled prediction matches a target population. However, this calibration is only
sound when the individual predictions are sound. Individual fidelity is
therefore the primary objective, and distributional alignment follows from it, not the other way around.

\paragraph{Ethics and data privacy.}
Anacreon learns from public text, and its design keeps the focus on groups rather
than named persons. The clustering step (\cref{sec:clustering}) is central to this.
Records are pooled into clusters, and every model is trained on a cluster rather
than on any single individual. The seed individuals only anchor the clusters. The
trained models capture behavioural trends shared across a cluster's members, not a
profile of any one person. Anacreon is therefore built to describe how a segment
tends to respond, not to identify, track, or manipulate a specific individual. Anacreon infers
general patterns from simulation, and its outputs are population-level tendencies
rather than predictions about private individuals.

\section{Conclusion}
\label{sec:conclusion}

This report set out to narrow the persistent gap between
prediction at the level of the aggregate and prediction at the level of the
individual (\cref{sec:introduction}). Statistical prediction has long paid out at
the level of the mass, in insurance or public health. It
has disappointed when pushed toward the specific person or act, in election forecasting. 
LLM-based simulators inherit this boundary. They tend to recover a population's central tendencies while
flattening the heterogeneity and minority positions that define the individuals
within it. Anacreon grounds a
language model in real qualitative data for a narrow, well-specified population,
then predicts how that population would answer questions it has never been asked.

The method makes several contributions (\cref{sec:method}). It builds a seed
population and an authorship embedding that separates individuals, clusters the
corpus around them, and trains a mixture of minds, one dedicated model per
cluster. It harvests
demographics, psychological traits, and survey responses from public text,
following the principle that public behaviour predicts private disposition. It also
confronts the two failure modes that the literature documents in naive LLM
simulation. Prompt brittleness is reduced by asking each question many times with
shuffled options. Bias is reduced by flipping questions between positive and
negative framing, which balances the responses the model sees during training.

These choices yield measurable gains. On a large, externally sourced survey
instrument, Anacreon reaches an ordinal alignment of $0.775$
(\cref{sec:main_results}). This is the state of the art among the systems that
report the measure. The residual positive bias is small and much reduced relative
to a base model, though it is not eliminated (\cref{sec:bias}). The weakest clusters
are not concentrated in any demographic group, which points to data quality within
a cluster, rather than the demographic it represents, as the limit on performance
(\cref{sec:analysis}).

For three and a half
centuries, prediction has paid out at the level of the mass and faltered at the
level of the person. Anacreon does not dissolve that boundary, but it does move it. By
grounding a mixture of minds in what individuals reveal about themselves, it
simulates a narrow population well enough to set the state of the art on
individual-level accuracy. Bias and variance remain, and population-level fidelity
is still to be shown.

\bibliographystyle{plainnat}
\bibliography{references}

\end{document}